\documentclass[letterpaper, 10 pt, conference]{ieeeconf}  

\IEEEoverridecommandlockouts                              

\usepackage{graphics} 
\usepackage{epsfig} 
\usepackage{times} 
\usepackage{amsmath} 
\usepackage{amssymb}  
\usepackage{booktabs}
\usepackage{multirow}
\usepackage{url}
\usepackage{hyperref}
\let\labelindent\relax
\usepackage[shortlabels]{enumitem}
\title{\LARGE \bf
SPIN Road Mapper: Extracting Roads from Aerial Images via Spatial and Interaction Space Graph Reasoning for Autonomous Driving
}

\author{Wele Gedara Chaminda Bandara, Jeya Maria Jose Valanarasu, and Vishal M. Patel
\thanks{Authors are with the Department of Electrical and Computer Engineering, The Johns Hopkins University, Baltimore, MD, USA. Emails: 
{\tt\small \{wbandar1, jvalana1, vpatel36\}@jhu.edu}.}%
}

\begin{document}

\maketitle
\thispagestyle{empty}
\pagestyle{empty}

\begin{abstract}
Road extraction is an essential step in building autonomous navigation systems. Detecting road segments is challenging as they are of varying widths, bifurcated throughout the image, and are often occluded by terrain, cloud, or other weather conditions. Using just convolution neural networks (ConvNets) for this problem is not effective as it is inefficient at capturing distant dependencies between road segments in the image which is essential to extract road connectivity. To this end, we propose a Spatial and Interaction Space Graph Reasoning (SPIN) module which when plugged into a ConvNet performs reasoning over graphs constructed on spatial and interaction spaces projected from the feature maps. Reasoning over spatial space extracts dependencies between different spatial regions and other contextual information. Reasoning over a projected interaction space helps in appropriate delineation of roads from other topographies present in the image. Thus, SPIN extracts long-range dependencies between road segments and effectively delineates roads from other semantics. We also introduce a SPIN pyramid which performs SPIN graph reasoning across multiple scales to extract multi-scale features. We propose a network based on stacked hourglass modules and SPIN pyramid for road segmentation which achieves better performance compared to existing methods. Moreover, our method is computationally efficient and significantly boosts the convergence speed during training, making it feasible for applying on large-scale high-resolution aerial images. Code available at: \url{https://github.com/wgcban/SPIN_RoadMapper.git}.
\end{abstract}

\section{INTRODUCTION}
Among all the topographic objects found in aerial images, road is one of the essential topographic features with numerous applications ranging from automatic navigation and guidance systems. Extraction of roads from aerial images helps to understand the connectivity between places and thus aid in automating navigation, disaster mitigation, and controlling traffic. Furthermore, road detection helps to determine the drivable areas for autonomous vehicles so that motion planning algorithms can be constrained on drivable roads. In addition, most of the algorithms designed for road boundary extraction and curb detection are based on road segmentation maps as the primary step \cite{mnih2010learning,mattyus2017deeproadmapper}.  The extraction of road boundaries and curbs can be used to further improve the safety of autonomous driving \cite{xu2021icurb,xu2021topo}. 

\begin{figure}[t]
	\centering
	\includegraphics[width=0.46\textwidth]{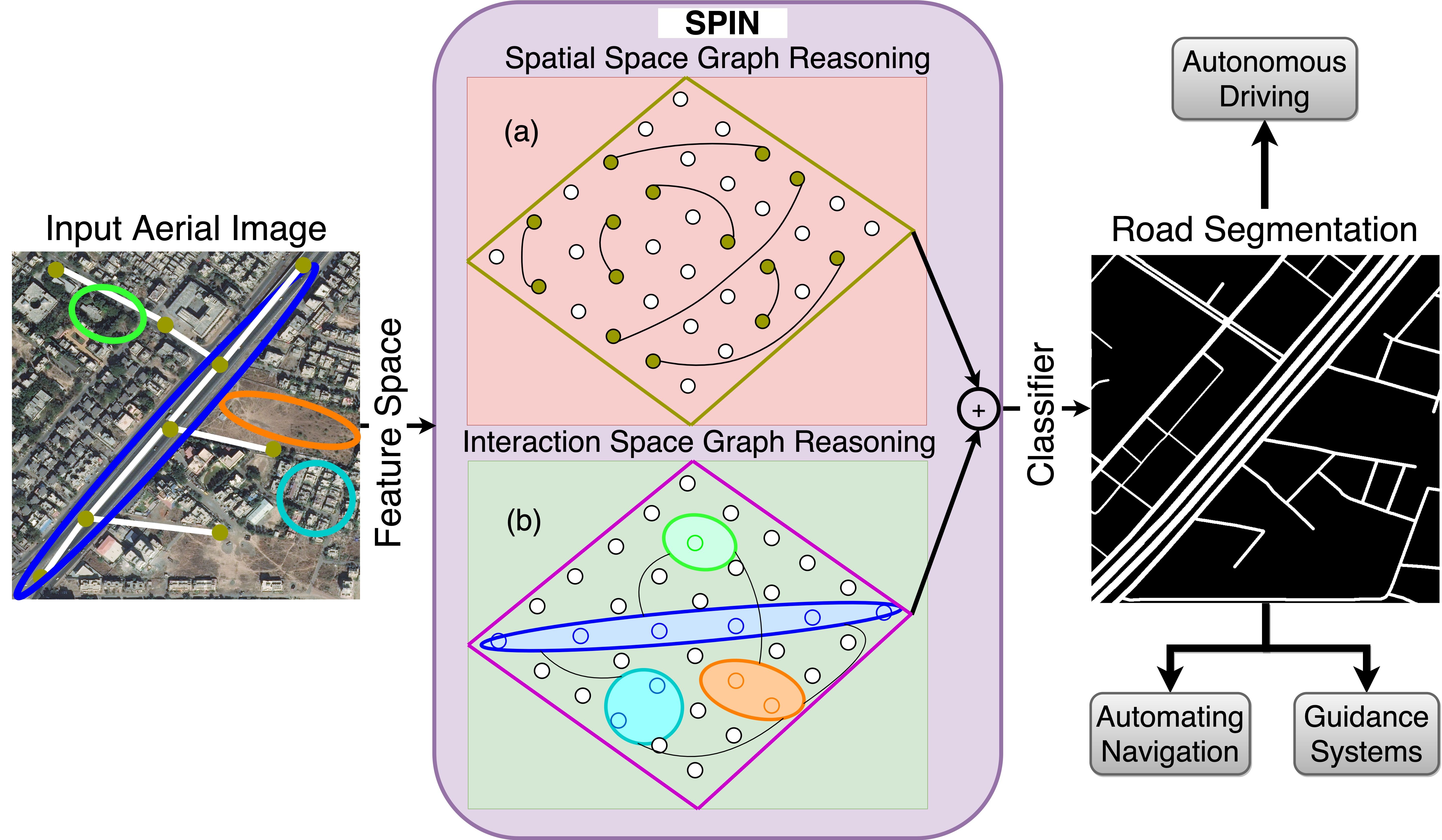}
    \vskip-10pt	
    \caption{An overview of our proposed method. We build graphs in two spaces: (a) spatial space and (b) a projected latent interaction space from feature maps. Graph reasoning in spatial space extracts connectivity between the road segments, whereas reasoning over interaction space delineates roads from other topographies. Nodes connected with lines in (a) denote how road segments are modeled to understand connectivity in the spatial space. Regions marked with different colors in (b) denote how different semantics are segregated for better road delineation in the interaction space.
	}
	\label{roadintro}
\end{figure}

Classical methods for road segmentation involve geometric-stochastic models \cite{barzohar1996automatic, vitor2014probabilistic}, line network extraction \cite{chai2013recovering}, and snakes \cite{laptev2000automatic}. There have also been works that consider the problem of road extraction as a problem of graph extraction from images \cite{hu2007road,hinz2003automatic}. Following the popularity of deep learning methods in computer vision \cite{krizhevsky2012imagenet,he2016deep}, techniques involving ConvNets have been explored for automatic road extraction \cite{mattyus2017deeproadmapper,mendes2016exploiting,costea2016aerial,bastani2018roadtracer}. These works pose road extraction as a problem of semantic segmentation where one tries to classify the pixels corresponding to the road from other semantics of the image.

Segmenting roads from aerial images is not a straight-forward segmentation problem because roads appear at different scales in the image due to varying widths and certain road regions are often narrow and get occluded with respect to the terrain. Also, there exists some similarity of the road texture with respect to nearby regions and there are chances of occlusion due to clouds and various weather conditions. One major problem of using ConvNets directly for road segmentation is that they are not good at learning long-range dependencies due to their inherent inductive biases. In aerial images, the road structure is mostly branched throughout the image as road is a connected topography. Also, just using a ConvNet does not constrain the network to learn representations of connected road segments \cite{mosinska2018beyond}. These issues make road segmentation from aerial images an open and challenging problem.

In this work, we focus on improving road segmentation by incorporating a global understanding of the image. Modeling dependencies and relations over regions in the image can help in understanding connectivity between the road segments. We note that transformer-based methods \cite{dosovitskiy2020image} are currently becoming popular for their property of extracting long-range dependencies. However, it is not feasible for applications on large-scale high-resolution remote sensing  datasets as it requires high compute power and significant training time. Thus, we propose using graph reasoning rather than just relying only on stacked convolutions or transformers to model global dependencies. Performing graph reasoning is light-weight and does not add on much to computation cost like transformers. 

A graph convolution \cite{kipf2016semi} can extract dependencies over distant regions making it more meaningful for using  it to understand road information on a global scale in aerial images. Graph convolutions have been explored for  video recognition \cite{wang2018videos}, semantic segmentation \cite{chen2019graph} and semi-supervised classification \cite{kipf2016semi}. Unlike these works, we propose performing graph reasoning in two domains - spatial and interaction space. In graph reasoning over spatial space, we build a graph over the feature space to extract dependencies between different spatial regions in the input. As we operate on the original coordinate space, performing reasoning over the graph would help to extract rich contextual information for road segmentation. For graph reasoning over interaction space, we construct a new interaction space where we model semantics with similar information together. This causes different semantic objects of the aerial image like roads, buildings, clouds, trees, and other topographic features to be modeled into different spaces. Performing graph reasoning over this interaction space would help in appropriate delineation of roads from other topographies in the image. Combining both, we propose a stand alone Spatial and Interaction space (SPIN) graph reasoning module which performs reasoning in the spatial and interaction space projected from the feature maps. Fig \ref{roadintro} illustrates how SPIN module helps to make road segmentation better. 

SPIN extracts long range dependencies between road segments and is effective at delineating roads from other semantics present in the image. When added to a base network, we show that it improves the segmentation performance by a reasonable amount.  It has numerous other advantages as well. SPIN can be plugged easily into a ConvNet architecture after a convolutional block. As SPIN learns highly contextual information, it increases the convergence rate of the network by half saving a lot of training time. This property is highly useful for training ConvNets on large-scale high-resolution images like aerial images. Adding SPIN to a ConvNet is also computationally effective as it adds on only $0.03 M$ parameters. Our proposed network consists of a feature extractor using residual blocks, stacked hourglass modules with skip connections for deep feature extraction and SPIN pyramid for graph reasoning. We analyze the effectiveness of our proposed method by conducting experiments on two large-scale road segmentation datasets - DeepGlobe \cite{demir2018deepglobe} and Massachusetts Road \cite{mnih2013machine} where we achieve a better performance than existing methods in the literature.

In summary, this paper makes the following contributions:
\begin{itemize}{}
	\item We propose a new module - Spatial and Interaction Space Graph Reasoning (SPIN), which when plugged into a ConvNet performs reasoning over graphs constructed on spatial and interaction space projected from the feature maps.
	\item We propose a new network built using stacked hourglass modules and SPIN pyramid for road segmentation from aerial images.
	\item We conduct extensive experiments on large-scale road segmentation datasets where we achieve better performance than existing methods both qualitatively and quantitatively.
	\item Our SPIN module is highly computationally efficient and helps in fast network convergence which makes training ConvNets on high-resolution aerial images quick and effective. 
\end{itemize}

\section{Related Work}
\noindent {\bf{Road segmentation:}}  Road segmentation is a well-studied problem in which we classify each pixel in a given aerial image as ``road" or ``no road" \cite{bastani2018roadtracer}. Early research on road segmentation primarily relied on probabilistic models to enhance connectivity by combining contextual prior conditions such as road geometry \cite{6185661, stoica2004gibbs} and color intensity \cite{WANG2016271}. In \cite{barzohar1996automatic}, geometric probability models were used to represent road images, and then maximum likelihood estimation (MLE) was used to predict road pixels. In \cite {wegner2013higher}, a model based on high-order conditional random fields (CRF) was used to incorporate prior knowledge of roads. However, these traditional probabilistic methods require hand-designed features and complex optimization techniques \cite{6185661}.

 One of the earliest attempts to automatically learn features for detecting roads in aerial images using expert labeled data was proposed in \cite{mnih2010learning}. In this study, unsupervised learning methods such as Principal Component Analysis (PCA) was used to initialize the feature detectors. Later, with the introduction of ConvNets in deep learning, researchers have investigated various ConvNet architectures to efficiently extract roads from aerial images \cite{abdollahi2020deep}. Among those, encoder-decoder based architectures are widely used due to its ability to capture relatively large spatial context \cite{abdollahi2020deep, bastani2018roadtracer}. Examples of these include U-Net \cite{ronneberger2015u}, LinkNet \cite{chaurasia2017linknet}, ResNet18 \cite{he2016deep} and multi-branch ConvNets \cite{batra2019improved, newell2016stacked}. In addition to the architectural changes, researchers also investigated different types of loss functions to replace well-known binary cross-entropy loss (BCE) to further improve the quality of road proposals and to incorporate topological constraints. In \cite{mattyus2017deeproadmapper}, a differentiable IoU loss function was proposed and most of the later work on road segmentation then used it instead of the BCE loss or combined them together for obtaining improved performance. Instead of just formulating the road extraction as a binary segmentation task, \cite{batra2019improved} introduced a multi-task learning \cite{kendall2018multi} approach where both segmentation and orientation of road line segments are used to improve the connectivity of the predicted road networks.\\   
\noindent {\bf{Graph convolutions:}}  The main limitation of Fully Convolutional Networks (FCNs) is its limited receptive field \cite{araujo2019computing}. To improve the receptive field of FCNs, researchers have proposed different solutions, such as adding pooling layers \cite{araujo2019computing}, dilated convolutions \cite{yu2015multi}, depth-wise convolutions \cite {howard2017mobilenets}, etc. However, these methods generally learn relations implicitly and are computationally expensive \cite{li2020spatial}. Instead, graph convolutions have the potential advantage of performing global reasoning on feature maps with explicit semantic meaning embedded in the graph structure. Due to this reason, many researchers have used graph convolutions in various computer vision tasks such as visual recognition \cite{li2018beyond, liang2018symbolic}, semantic segmentation \cite{chen2019graph, li2020spatial, li2019global} and  semi-supervised classification \cite{kipf2016semi}. In \cite{li2020spatial}, a graph reasoning module was proposed to capture multiple long-range contextual patterns of the original feature map through a data-dependent similarity matrix. In contrast to \cite{li2020spatial}, \cite{chen2019graph}  first  transformed  the original feature space into another latent coordinate space called interaction space and performed relational reasoning via graph convolution in the interaction space. In our proposed SPIN module, we combine the reasoning power of both spatial and interaction space graph reasoning by concatenating the individual outcomes. Further, we perform SPIN graph reasoning on different scales of the feature maps to learn multi-scale contextual relationships.

\begin{figure}[tb!]
	\centering
	\includegraphics[width=\linewidth]{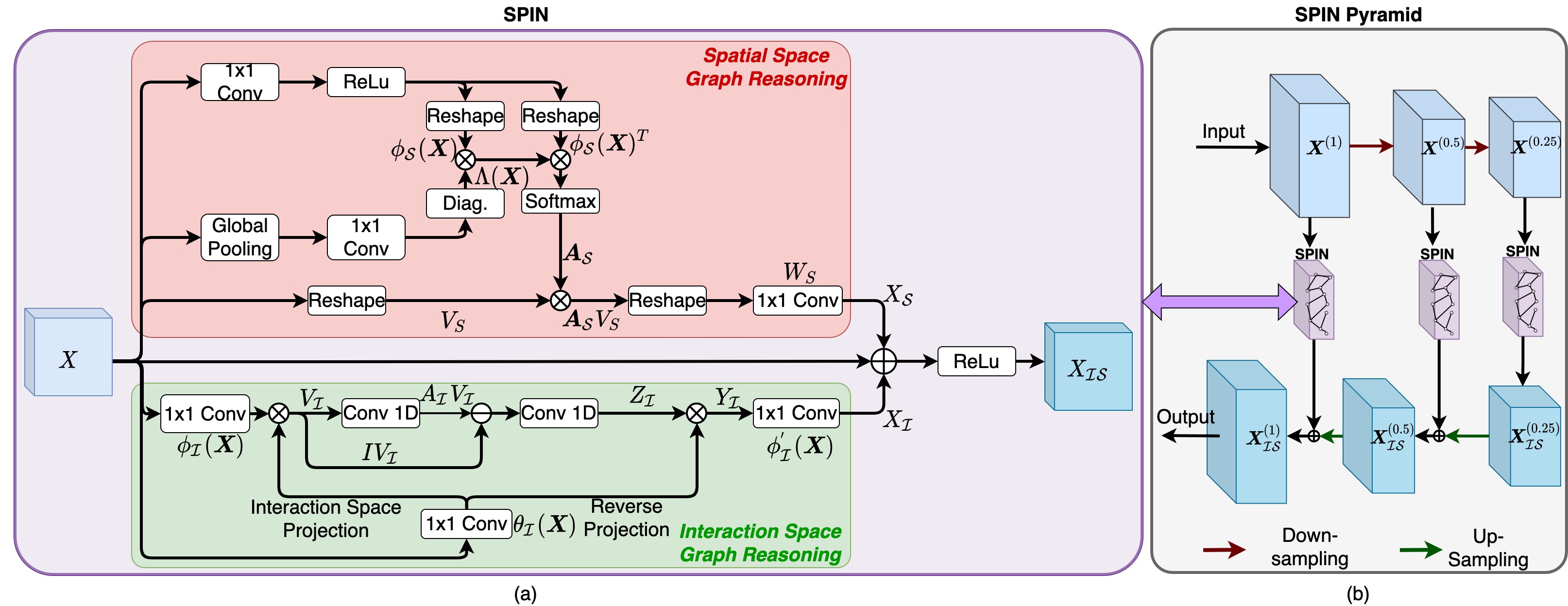}
    \vskip-10pt
    \caption{The architecture of our proposed method. (a) We perform graph reasoning in both spatial and interaction space. (b) The proposed SPIN pyramid module which performs SPIN graph reasoning at multiple scales ($1, 1/2, \text{ and } 1/4$) of original feature map to extract multi-scale long-range contextual information.
	\vspace{-5mm}
	}
	\label{SPIN}
\end{figure}
\section{Proposed Method}
\subsection{SPIN graph reasoning}
The proposed SPIN module is shown in Figure \ref{SPIN}-(a). We build
graphs in two spaces: spatial space and a projected latent interaction space from input feature maps. Then, graph reasoning is performed in spatial space to improve the connectivity between road segments and interaction space to delineate roads from other topographies. Assuming that spatial and interaction space graph reasoning provide different feature representations, we concatenate the output feature maps of both graph reasoning modules, as shown in Figure \ref{SPIN}-(a), to extract rich global contextual information of road segments. In order to capture multi-scale context of input feature maps, we build SPIN pyramid by performing SPIN graph reasoning on different scales and then aggregate them  as shown in Figure \ref{SPIN}-(b). 

In what follows, we elaborate on each block of the SPIN module in detail.  Before that, we briefly review graph reasoning.\\
\noindent {\bf{Graph reasoning:}}  A graph $G = (\boldsymbol{\mathcal{V}}, \boldsymbol{\mathcal{E}}, \boldsymbol{A})$ is defined by its nodes $\boldsymbol{\mathcal{V}}$, edges $\boldsymbol{\mathcal{E}}$ and similarity matrix $\boldsymbol{A}$ that describes the similarity between each and every pixel (node) in the graph. Let $\boldsymbol{X} \in \mathbb{R}^{L \times C}$ denote the input feature map where  $C$ is the number of channels and $L = W \times H$.  Here, $W$ and $H$ correspond to the width and height of $\boldsymbol{X}$. Standard 2D convolutions only share information among the positions in a small neighborhood defined by the filter size. In order to achieve a large receptive field and to capture long-range dependencies among the pixels, ConvNet architectures stack multiple convolution layers which is highly inefficient. In contrast, a single graph convolution layer can extract long-range dependencies of input feature map very efficiently and effectively. Formally, the graph convolution is defined as \cite{kipf2016semi},
\setlength{\belowdisplayskip}{0pt} \setlength{\belowdisplayshortskip}{0pt}
\setlength{\abovedisplayskip}{0pt} \setlength{\abovedisplayshortskip}{0pt}
\begin{equation}
	\boldsymbol{\tilde{X}} = \sigma \left( \boldsymbol{A} \boldsymbol{X} \boldsymbol{W} \right),
	\label{def_GR}
\end{equation}
where $\boldsymbol{W}$ is the learnable weight matrix (usually modeled as a convolutional layer), $\sigma(\cdot)$ is the non-linear activation function (e.g. ReLU) and, $\boldsymbol{A}$ and $\boldsymbol{X}$ are the same as defined above. Note that the only difference between graph convolution and conventional convolution is that in graph convolution, we left-multiply the original feature map  $\boldsymbol{X}$ by the similarity matrix  $\boldsymbol{A}$ before doing the convolution operation. 

With this background of graph reasoning, we now describe the two main building blocks of our proposed SPIN module: (1) Spatial space graph reasoning and, (2) Interaction space graph reasoning.

\subsubsection{Spatial space graph reasoning}
The overall procedure of spatial space graph reasoning is depicted in the red box in Figure \ref{SPIN}-(a). As described in the previous section, the main intuition behind spatial space graph reasoning is to improve the connectivity between the predicted road segments. We first build a fully-connected graph in the spatial domain $\mathcal{S}$ using the spatial similarity matrix $\boldsymbol{A}_{\mathcal{S}}$ and then perform spatial graph reasoning. We now describe the computation procedure of spatial graph reasoning in detail.\\
\noindent {\bf{Computation of spatial similarity matrix $\boldsymbol{A}_{\mathcal{S}}$:}}  The first step of spatial graph reasoning is to compute the spatial similarity matrix $\boldsymbol{A}_{\mathcal{S}} \in \mathcal{R}^{L \times L}$. There are different similarity metrics that have been proposed in the literature to calculate the similarity between two given pixels. The most popular are the Euclidean distance and the dot product. In our implementation, we use the dot product to compute the similarity matrix $ \boldsymbol {A}_{\mathcal{S}} $.

The similarity matrix $\boldsymbol{A}_{\mathcal{S}}$ for an input feature map $\boldsymbol{X}$ can be represented as a multiplication of three transformations as follows:
\begin{equation}
	\boldsymbol{A}_{\mathcal{S}} = \text{Softmax}( \phi_{\mathcal{S}}(\boldsymbol{X}) \Lambda(\boldsymbol{X}) \phi_{\mathcal{S}}(\boldsymbol{X})^{T}),
	\label{similarity}
\end{equation}
where $\phi_{\mathcal{S}}(\boldsymbol{X}) \in \mathcal{R}^{L \times M}$ is a linear transformation followed by ReLU non-linearity and $\Lambda(\boldsymbol{X}) \in \mathcal{R}^{M \times M}$ is the diagonal matrix. Note that $M$ is the dimension of the intermediate feature map.

In this implementation, the linear transformation $ \phi_{\mathcal{S}}(\boldsymbol{X})$ is modeled using a $1 \times 1$   convolution layer that reduces input feature map dimension from $C$ to $M$. The transformation $\Lambda(\boldsymbol{X})$ is represented by a global average pooling followed by a $1 \times 1$ convolution. Then we reshape the outputs $\phi_{\mathcal{S}}(\boldsymbol{X})$ and $\Lambda(\boldsymbol{X})$ appropriately to perform matrix multiplication as shown in Figure \ref{SPIN}-(a) to obtain the similarity matrix $\boldsymbol{A}_{\mathcal{S}} \in \mathcal{R}^{L \times L}$.\\
\noindent {\bf{Graph reasoning in spatial space:}} Once we have the similarity matrix $\boldsymbol{A}_{\mathcal{S}}$, we can perform the spatial graph reasoning on input data according to the Eq. \eqref{def_GR}. First, we reshape the input data appropriately and then we perform the matrix multiplication to obtain $\boldsymbol{A}_{\mathcal{S}} \boldsymbol{X}$. Next, we multiply it by the trainable weight matrix $\boldsymbol{W}_{\mathcal{S}}$ that is modeled as a $1 \times 1$ convolution layer as shown in Figure \ref{SPIN}-(a). Finally, we apply ReLU to obtain the spatial graph reasoned feature matrix $\boldsymbol{X}_{\mathcal{S}}$.
\subsubsection{Interaction space graph reasoning}
The overall procedure of interaction space graph reasoning is shown in the green box in Figure \ref{SPIN}-(a). As we described earlier, the spatial space graph reasoning can improve the connectivity between predicted road segments. We now consider projecting the input feature space into another latent space, called the interaction space $ \mathcal{I} $, where we try to delineate roads from other objects such as buildings, trees, vehicles, etc. Next, we build a graph that connects these features in the interaction space and performs a relational reasoning on the graph. After reasoning, the updated information is projected back to the original coordinate space. In what follows, we discuss these operations in detail.\\
\noindent {\bf{Projection to interaction space:}}  The first step is to project the original feature map $\boldsymbol{X}$ to the interaction space $\mathcal{I}$. This is done by the projection function $f(\cdot)$ such that the features $\boldsymbol{V}_{\mathcal{I}} \in \mathcal{R}^{N \times S}$ in the interaction space are more friendly for global reasoning over disjoint and distant regions, where $N$ is the number of nodes and $S$ is the number of states.

In practice, we first reduce the dimension of the input feature $\boldsymbol{X}$ with the transformation $\theta_{\mathcal{I}}(\boldsymbol{X}) \in \mathcal{R}^{L \times N}$ and formulate the projection function $\phi_{\mathcal{I}}(\boldsymbol{X}) \in \mathcal{R}^{L \times S}$ as a linear combination of input $\boldsymbol{X}$ such that the new features can aggregate information from multiple regions. Concretely, the input feature $\boldsymbol{X}$ is projected as $V_{\mathcal{I}}$ in the interaction space $\mathcal{I}$ through the projection function $\phi_{\mathcal{I}}(\boldsymbol{X})$ as follows:
\begin{equation}
	\boldsymbol{V}_{\mathcal{I}} = \theta_{\mathcal{I}}(\boldsymbol{X})^T \phi_{\mathcal{I}}(\boldsymbol{X}).
\end{equation}
We implement both functions  $\phi_{\mathcal{I}}(\cdot)$ and $\theta_{\mathcal{I}}(\cdot)$ as $1 \times 1$ convolutional layer as shown in Figure \ref{SPIN}-(a).\\
\noindent {\bf{Graph reasoning in interaction space:}} After projecting the input feature space into interaction space, we build a fully-connected graph in the interaction space with the node similarity matrix $\boldsymbol{A}_{\mathcal{I}} \in \mathcal{R}^{N \times N}$. The similarity matrix $\boldsymbol{A}_{\mathcal{I}}$ is randomly  initialized and learned during back propagation in contrast to the similarity matrix we defined for the spatial graph reasoning that is dependent on the input data. In addition, we use skip connection (i.e. identity matrix) that speeds up the optimization. Following Eq. \eqref{def_GR}, the graph convolution in the interaction space is formulated as:
\begin{equation}
	\boldsymbol{Z}_{\mathcal{I}} = \boldsymbol{A} \boldsymbol{X} \boldsymbol{W} = ((\boldsymbol{I}-\boldsymbol{A}_{\mathcal{I}}) \boldsymbol{V}_{\mathcal{I}})\boldsymbol{W}_{\mathcal{I}},
\end{equation}
where $\boldsymbol{W}_{\mathcal{I}}$ is the trainable weight matrix. Here both matrices  $\boldsymbol{W}_{\mathcal{I}}$ and $\boldsymbol{A}_{\mathcal{I}}$ are implemented as 1D convolution with kernel size of 1 as shown in Figure \ref{SPIN}-(a).\\
\noindent {\bf{Reverse projection to the original coordinate space:}} After graph reasoning in the interaction space, we project the output features $\boldsymbol{Z}_{\mathcal{I}}$ to the original coordinate as:
\begin{align}
	\boldsymbol{Y}_{\mathcal{I}} &= \theta_{\mathcal{I}}(\boldsymbol{X})^T\boldsymbol{Z}_{\mathcal{I}},\\
	\boldsymbol{X}_{\mathcal{I}} &= \phi^{'}_{\mathcal{I}}(\boldsymbol{Y}_{\mathcal{I}}).
\end{align}
 We use the same projection matrix $\theta(\boldsymbol{X})$ to transform features to $\boldsymbol{Y}_{\mathcal{I}} \in \mathcal{R}^{L \times S}$. Then we perform linear projection $\phi^{'}_{\mathcal{I}}(\cdot)$ using a $1 \times 1$ convolution layer to transform $\boldsymbol{Y}_{\mathcal{I}}$ into the original coordinate space. As a result we have the features $\boldsymbol{X}_{\mathcal{I}}$ with feature dimension $C$ at the original coordinate space. 
 
 Once we have the spatial and interaction space graph reasoning outputs, we combine them with the original input feature map and then apply ReLU non-linearity to get the final graph reasoned feature map $\boldsymbol{X}_{\mathcal{I}\mathcal{S}}$. Mathematically, we can denote this as,
 \begin{equation}
 \boldsymbol{X}_{\mathcal{I}\mathcal{S}} = \text{ReLU} (\boldsymbol{X}_{\mathcal{S}}+\boldsymbol{X}+\boldsymbol{X}_{\mathcal{I}}).
 \end{equation}

\subsubsection{SPIN pyramid}We perform our SPIN graph reasoning at multiple scales to further increase the overall receptive field of the network and to improve long-range contextual information present in the intermediate feature maps. Concretely, we perform SPIN graph reasoning at different scales ($1, 1/2, \text{ and } 1/4$) of the original feature map as shown in Figure \ref{SPIN}-(b). In the results and discussion section, we conduct an ablation study to demonstrate the effect of spatial, interaction, and SPIN graph reasoning on the segmentation performance.

\begin{figure}[t!]
	\centering
	\includegraphics[width=\linewidth]{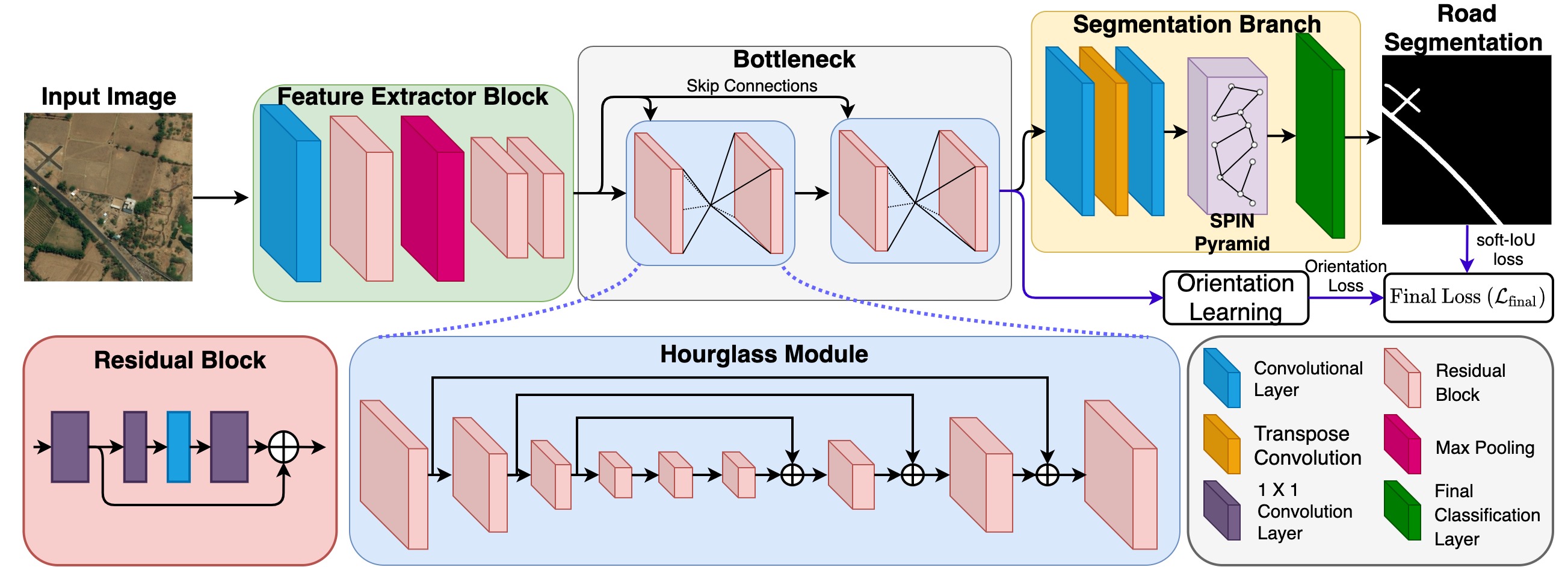}
    \vskip-10pt	
    \caption{Proposed network for road segmentation from aerial images. 
	}
	\label{Proposed_network}
	\vspace{-5mm}
\end{figure}

\subsection{Network architecture}
\noindent {\bf{Feature extractor block:}} Operating the network at high-resolution (i.e. $256\times256$) requires a large GPU memory and computational power. Therefore, we employ a $7 \times 7$ convolutional layer with stride 2, followed by a residual block and a max pooling layer to bring it down to the size of $64 \times 64$. We then add two subsequent residual modules before sending it to the hourglass module.\\
\noindent {\bf{Bottleneck:}} Our proposed road segmentation network uses stack of two hourglass modules \cite{newell2016stacked} at the bottleneck. The hourglass module captures information at different scales by cascading series of residual modules and max pooling layers. When the network reaches the lowest resolution, it performs bilinear upsampling and combines features across the same scales using skip connections. We feed forward the output of the bottleneck to the segmentation branch.\\
\noindent {\bf{Segmentation branch:}} In the segmentation branch, we use a combination of convolution and transpose convolution layers to upsample the feature maps to the original scale. We then feed forward these feature maps to our proposed SPIN pyramid. To get the output segmentation map, we feed forward these graph reasoned feature maps from the SPIN pyramid to the final classification layer.\\  
\noindent {\bf{Orientation learning:}} For orientation learning, we adopt the same orientation learning technique described in Batra \textit{et. al.} \cite{batra2019improved}. As shown in Figure \ref{Proposed_network}, our road segmentation network is divided into two branches after the two hourglass modules to support for both segmentation and orientation learning. The orientation learning task is formulated as a multi-class classification problem where, the orientation of each road-pixel is quantized into bins resulting in a total of 37 orientation classes. Please check the supplementary document for more details.\\
\noindent {\bf{Loss function:}} The proposed road segmentation network utilizes predictions from intermediate feature maps to compute the loss at multiple scales: $((H/4,W/4)$, $(H/2 , W/2)$ and $(H , W))$ instead of computing it only at the original scale. This improves network's ability to correctly predict road segments at multiple scales and helps to convergence faster. In this implementation, we make use of two loss functions: (1) segmentation loss, and (2) orientation loss. We use differentiable SoftIoU loss to compute the segmentation loss instead of using the BCE loss. The  segmentation loss is computed at multiple scales.  The road segmentation loss $\mathcal{L}_{\text{seg}}$ is defined as follows,
\begin{equation}
	\mathcal{L}_{\text{seg}}  =  \sum_{s} \left(1 - \mathrm{SoftIoU}(Y_{\text{pred}}^{s}, Y_{\text{gt}}^{s}) \right),
\end{equation}
where $s$ denotes the scale having values $\left\{ (H,W), (H/2,W/2), (H/4, W/4) \right\}$, $Y_{\text{pred}}^{s}$ and $Y_{\text{gt}}^{s}$ are the predicted and ground-truth segmentation maps at scale $s$, respectively. Similarly, we calculate the orientation loss at multiple scales. The orientation loss $\mathcal{L}_{\text{orient}}$ is defined as follows,
\begin{equation}
	\mathcal{L}_{\text{orient}} =  \sum_{s} \left(1 - \sum_{b=0}^{N_{\text{bins}}} O_b^s \log(\hat{O}_b^s)\right),
\end{equation}
where $N_{\text{bins}}$ is the number of bins in the quantized orientation, $O_{b}^{s}$ and $\hat{O}_{b}^{s}$  are the predicted and ground-truth orientation maps of orientation bin $b$ and scale $s$, respectively. Finally, the overall loss  function $\mathcal{L}$ is defined as follows,
\begin{equation}
	\mathcal{L}_{\text{final}} = \mathcal{L}_{\text{seg}} + \mathcal{L}_{\text{orient}}.
\end{equation}
\section{Experimental Settings}
\subsection{Datasets}
\noindent {\bf{Massachusetts road dataset:}} The Massachusetts Roads dataset~\cite{mnih2013machine} consists of train, validation and test sets with 1108, 14 and 49 images, respectively, each with a size of $ 1,500 \times 1,500$ pixels. Following \cite{wulamu2019multiscale}, we fill the training images into size of $1536 \times 1536$ and then we crop each image into $512 \times 512$ patches with overlapping window of $256$ pixels to make the training set. We observed that some parts of the images in the Massachusetts Road dataset are partially occluded and these images severely affect the performance of models. Hence, we removed these occluded images from the training set. Similarly, we crop each image in validation and test sets into $512 \times 512$ patches without any overlapping window. After these series of operations, the processed Massachusetts Road dataset contains $21782$, $124$, and $433$ images with size of $512 \times 512$, corresponding to the train, validation and test set, respectively.\\
\noindent {\bf{DeepGlobe dataset:}} For the DeepGlobe dataset ~\cite{demir2018deepglobe}, we follow the same experimental and data preparation protocols mentioned in \cite{batra2019improved}. The DeepGlobe dataset consists  of $6226$ images with resolution of $1024 \times 1024$. Following \cite{batra2019improved}, we create splits of $4696$ images for training and $1530$ images for testing. Then, we create the patches with $512 \times 512$ resolution with an overlapping window of 256 pixels and this results in total of $42264$ images for training. Similarly, for the testing dataset also we create patches with resolution of $512 \times 512$ without any overlapping pixels and this results in total of $6116$ images.\\

\begin{table*}[tbh!]
	\centering
	\caption{A quantitative comparison of our SPIN Road Mapper with the SOTA baselines in terms of F1 score, $IoU^r$ and $IoU^a$.}
	\begin{tabular}{lp{8mm}p{6mm}p{5mm}p{5mm}p{5mm}p{5mm}cp{8mm}p{6mm}p{5mm}p{5mm}p{5mm}p{5mm}} %
		\toprule
		\multirow{2}{*}{\parbox[c]{.2\linewidth}{\centering Method}} & \multicolumn{6}{c}{Massachusetts Road Dataset \cite{mnih2013machine}} & & \multicolumn{6}{c}{DeepGlobe Dataset \cite{demir2018deepglobe}} \\ 
		\cmidrule{2-7} \cmidrule{9-14}
		
		& {Precision} & {Recall} & {F1} & {$IoU^r$} & {$IoU^a$} & APLS & & {Precision} & {Recall} & {F1} & {$IoU^r$} & {$IoU^a$} & APLS\\
		\midrule
		Seg-Net \cite{badrinarayanan2017segnet}			& 77.34 	& 79.84 & 78.57 & 64.71 & 58.59 & 57.76 && 69.48 & 72.97 & 71.19 & 55.26 & 49.20 & 58.55\\
		U-Net \cite{ronneberger2015u}					& 82.46 	& 84.34 & 83.39 & 71.51 & 60.97 & 61.33 && 73.55 & 74.98 & 74.26 & 59.06 & 55.02 & 61.23\\
		LinkNet \cite{chaurasia2017linknet}			& 83.25  	& 84.63 & 83.93 & 72.32 & 63.12 & 66.62 && 78.34 & 78.85 & 78.59 & 64.73 & 62.75 & 67.41\\
		HourGlass \cite{newell2016stacked}				& 81.26 	& 81.86 & 81.56 & 68.86 & 61.37 & 65.37 && 79.43 & 80.14 & 79.78 & 66.34 & 60.71 & 65.33\\
		Stack-HourGlass \cite{newell2016stacked}		& 80.12 	& 83.87 & 81.96 & 69.43 & 62.21 & 67.89 && 79.33 & 79.99 & 79.66 & 66.19 & 62.06 & 69.02\\
		Batra \textit{et al.} \cite{batra2019improved}	& 83.34 	& 84.61 & 83.97 & 72.37 & 64.44 & 71.34 && 83.79 & 84.14 & 83.97 & 72.37 & \textbf{67.21} & 73.12\\
		SPIN Road Mapper (ours) 						& \textbf{83.90} & \textbf{85.06} & \textbf{84.47} & \textbf{73.12} & \textbf{65.24} & \textbf{72.49} && \textbf{84.14} & \textbf{84.50} & \textbf{84.32} & \textbf{72.89} & 67.02 & \textbf{74.14}\\
		\bottomrule
	\end{tabular}
	\normalsize
	\label{quantitative_results}
\end{table*} 
\vspace{-10pt}
\subsection{Implementation details} We use random crops of resolution $256 \times 256$ to train the network for the Massachusetts and DeepGlobe datasets. We use extensive data augmentation techniques such as image rotation, flipping, and mirroring. We use SGD optimizer with a batch size of $32$, a momentum of $0.9$ and a weight decaying of $ 0.0005$. We use a step-learning rate scheduler with an initial learning rate of $0.01$ where steps are scheduled at $50$, $90$, and $110$. We reduce the learning rate by a factor of $0.1$ at each step. We train the network for a total of $120$ epochs. We implemented our model in PyTorch and used an NVIDIA Quadro RTX 8000 GPU for all of our experiments.\\

\vspace{-10pt}
\subsection{Performance metrics} For the DeepGlobe dataset accurate road segmentation masks are available and hence, we evaluate the quality of our road predictions using accurate road Intersection over Union ($IoU^a$) and $F1$ score. However, the groundtruth segmentation masks of Massachusetts road dataset have constant width and this will adversely affect the pure pixel based metrics. So, as proposed in \cite{mnih2012learning} we also use relaxed IoU ($IoU^r$) with buffer size of $4$ in our evaluations. Furthermore, we use Average Path Length Similarity (APLS) metric to measure the difference between ground truth and proposal graphs \cite{van2017spacenet}.

\section{Results}
\begin{figure}[tb]
	\centering
	\includegraphics[width=0.99\linewidth]{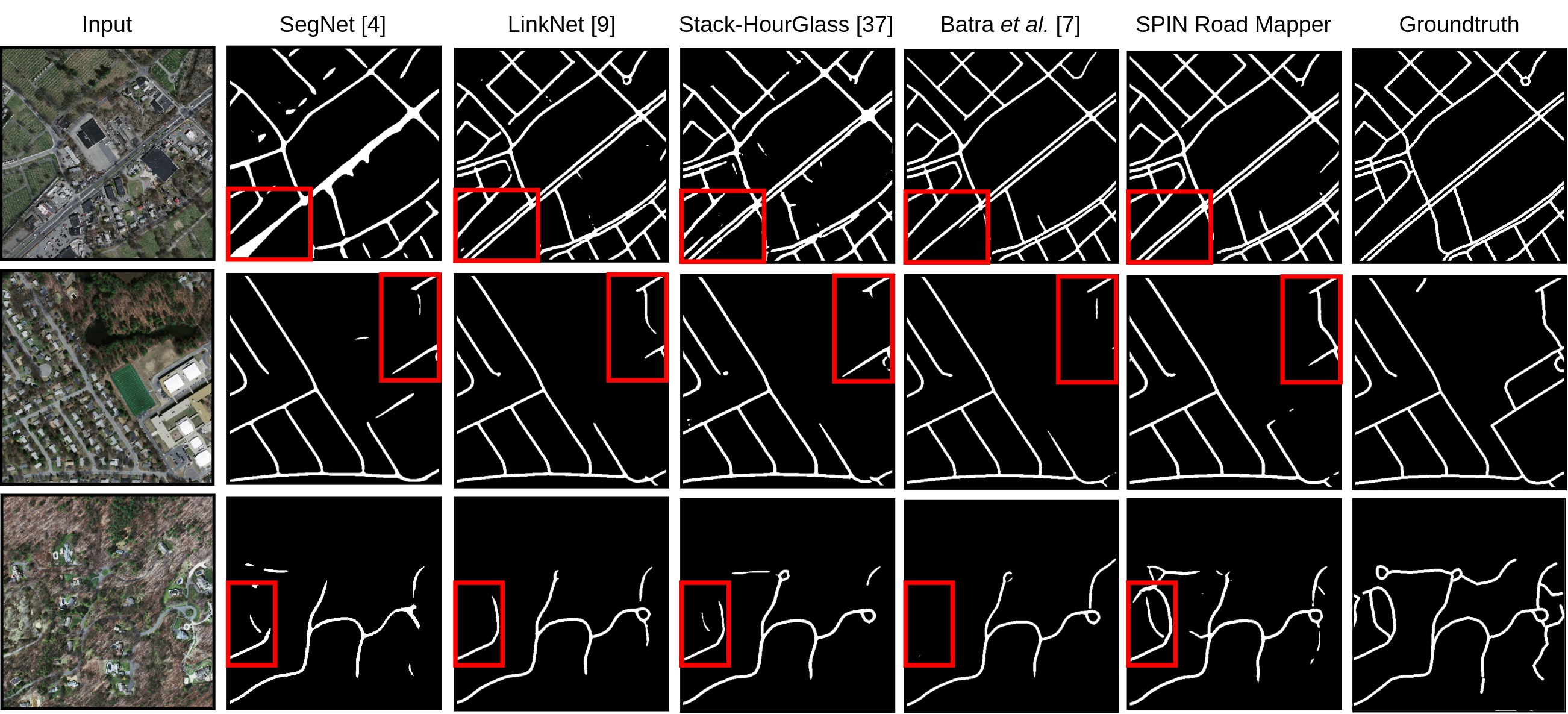}
    \vskip-20pt	=
    \caption{A qualitative comparison between our SPIN Road Mapper and the SOTA methods. 
	}
	\label{qualitative_results}
	\vspace{-5mm}
\end{figure}

\begin{figure}[tb]
	\centering
	\includegraphics[width=0.99\linewidth]{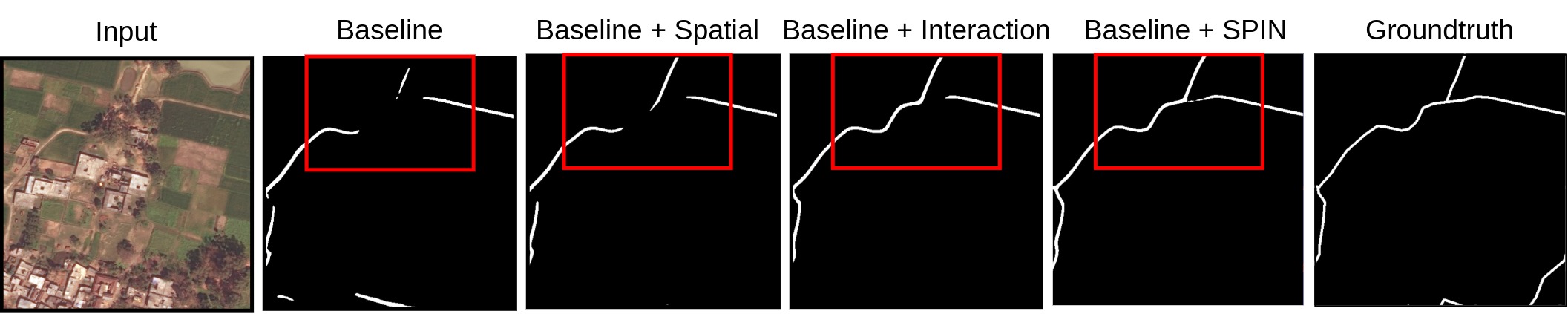}
    \vskip-10pt
    \caption{A qualitative comparison for the ablation study.
	}
	\label{abalation_qualitative}
\end{figure}

In this section, we compare the road segmentation performance of our SPIN Road Mapper with existing methods, quantitatively and qualitatively.  In particular, we compare the performance of our method with that of Seg-Net \cite{badrinarayanan2017segnet}, U-Net \cite{ronneberger2015u}, LinkNet34 \cite{chaurasia2017linknet}, HourGlass \cite{newell2016stacked}, Stack-HourGlass \cite{newell2016stacked}, and Batra \textit{et al.} \cite{batra2019improved}.\\
\noindent {\bf{Quantitative results:}} The quantitative results are summarized in Table \ref{quantitative_results}. As can be seen from Table \ref{quantitative_results}, the proposed SPIN Road Mapper achieves the state-of-the-art (SOTA) results in terms of all the performance measures for the Massachusetts dataset. When considering the DeepGlobe dataset, our method achieves the SOTA results in terms of F1, $IoU^r$, and APLS. Furthermore, the improvement in terms of APLS metric is significant (+1.15\% and +1.02\% for Massachusetts and DeepGlobe datasets, respectively) and which confirms that the proposed SPIN module improves the connectivity of road segments by specially bringing gaps for occluded areas (see Fig. \ref{abalation_qualitative}). These qualitative results confirms that the proposed SPIN module effectively captures the long-range dependencies of feature maps, and thereby helps final classifier to identify road pixels substantially well compared to the available ConvNet architectures for road segmentation.\\
\noindent{\bf{Qualitative results:}} For the qualitative analysis, we visualize the predicted road maps from SegNet \cite{badrinarayanan2017segnet}, LinkNet \cite{chaurasia2017linknet}, Stack-HourGlass \cite{demir2018deepglobe}, Batra \textit{et. al} \cite{batra2019improved}, and our SPIN Road Mapper on the Massachusetts Road dataset in Figure \ref{qualitative_results}. The red boxes in Figure \ref{qualitative_results} highlight the regions where our method performs better than the baseline methods. For example, consider the last row of Figure \ref{qualitative_results}. Roads in the region highlighted by the red box are mostly covered by trees and buildings (as can be seen from the input aerial image), making it difficult for the baseline segmentation networks to correctly identify the presence of roads. In contrast, our method is able to predict most of the road segments due to its ability to capture long-range dependencies between road pixels through spatial graph reasoning, as well as its ability to delineate roads from surrounding structures through interaction space graph reasoning.\\
\noindent {\bf{\hspace{-1pt}Ablation study:}} We conduct an ablation study to demonstrate the effect of spatial, interaction and SPIN graph reasoning on road segmentation. It can be seen from Table \ref{Abalation_study_GR} that integrating spatial and interaction space graph reasoning to the ConvNet-based network results in increase road segmentation accuracy. Combining the spatial and interaction space graph reasoning together in SPIN pyramid results in further improvement over the individual components. In addition to the quantitative comparison, we also present a qualitative comparison in Figure \ref{abalation_qualitative} which clearly demonstrates how each graph reasoning technique improves the quality of road predictions. These experiments show that out proposed SPIN pyramid helps the network learn features with more global contextual information resulting in an improved performance.

\begin{figure}[tb!]
	\vspace{-9pt}
	\centering
	\includegraphics[trim={15pt 0pt 15pt 10pt}, width=0.38\linewidth]{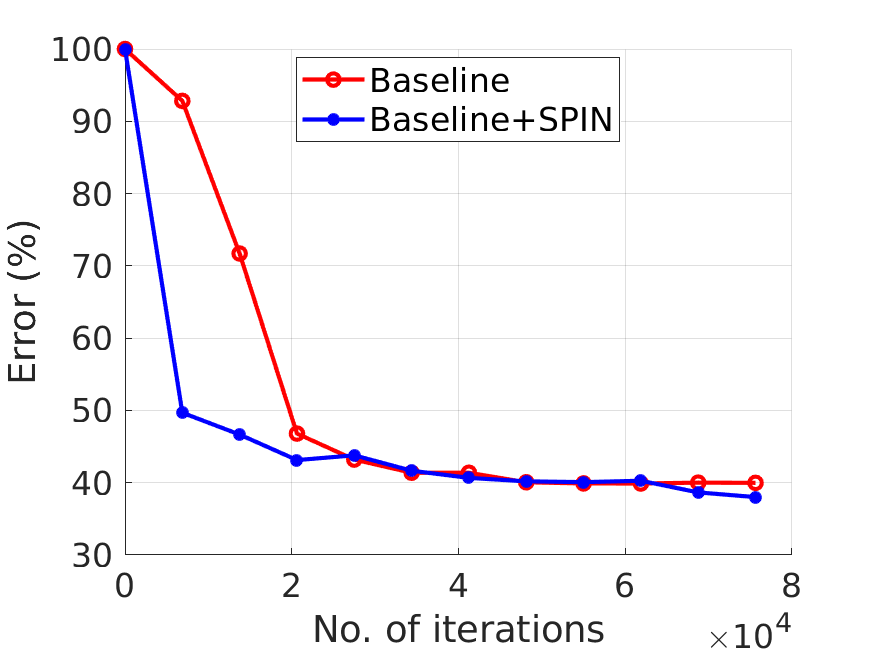}
    \vskip-10pt	
    \caption{The convergence characteristic for with and without the SPIN.}
	\label{convergence}
	\vspace{-5mm}
\end{figure}

\begin{table}[tb]
	\centering
	\caption{Quantitative results of ablation study.}
	\vskip-10pt
	\begin{tabular}{lccc} %
		\toprule
		Method & {$IoU^a$} & {F1} & APLS  \\
		\midrule
		ConvNet Only 	            & 83.97 & 66.58 & 73.01\\
		ConvNet + Spatial GR        & 84.13	& 66.82 & 73.59\\
		ConvNet + Interaction GR    & 84.12 & 66.76 & 73.52\\
		ConvNet + SPIN GR           & \textbf{84.32} & \textbf{67.02} & \textbf{74.14}\\
		\bottomrule
	\end{tabular}
	\label{Abalation_study_GR}
\end{table}
\noindent {\bf{Convergence:}} Figure \ref{convergence} shows the training convergence plot of the proposed network with and without the SPIN module. We can observe that adding the SPIN module helps achieve faster convergence. This leads to reduction in training time which is crucial for training ConvNets on large-scale high-resolution remote sensing datasets.

\section{Conclusion}
We presented a Spatial and Interaction Space Graph Reasoning (SPIN) module that can be plugged into ConvNets to learn distant relationships between road segments in the feature space. Learning global dependencies are essential while extracting complex road topology from  aerial images where most of the road segments are partially or completely occluded by trees, buildings, or clouds. The graph reasoning over the spatial space helps the network to extract more dependencies between different spatial regions and other contextual information whereas graph reasoning
over a projected interaction space helps to delineate roads from surrounding objects. We conduct extensive experiments and compare the predicted road maps qualitatively and quantitatively with existing methods. We observe that our SPIN module helps convolutional networks to extract long-range dependencies and thereby improve the segmentation quality. SPIN is computationally light and also helps in faster convergence which are crucial while training ConvNets on large-scale high-resolutions datasets.

\bibliographystyle{IEEEtran}
\bibliography{root}

\end{document}